\pdfoutput=1

\documentclass[11pt]{article}

\usepackage[]{emnlp2021}
\usepackage{times}
\usepackage{latexsym}
\usepackage[T1]{fontenc}
\usepackage[utf8]{inputenc}
\usepackage{microtype}

\usepackage{graphicx}
\usepackage{tabularx}
\usepackage{booktabs}
\usepackage{soul}
\usepackage{multirow}
\usepackage{color, colortbl}
\usepackage{enumerate}
\usepackage{tikz}
\usepackage{enumitem}
\usepackage{mdframed}
\usepackage{framed}
\usepackage{lipsum}
\usepackage{xstring}
\usepackage{xcolor}
\usepackage{float}
\usepackage{amssymb}
\usepackage{algorithmic}
\usepackage{listings}
\usepackage{hanging}
\usepackage{ifthen}
\usepackage{amsmath}
\usepackage{textcomp}
\usepackage{nameref} 
\usepackage{subcaption} 
\usepackage{hyperref} 
\hypersetup{ 
   colorlinks=true,
   citecolor=blue,
   linkcolor=blue,
   urlcolor=blue
}
\usepackage[colorinlistoftodos,prependcaption,textsize=tiny]{todonotes}

\newcommand{\exampletext}[1]{
    \hspace{1mm}
    \begin{minipage}[c]{1.0\linewidth}
    \begin{framed}
        \small \textit{#1}
    \end{framed}
    \end{minipage}
}
\newcommand{\ICNALEargs}{\mbox{ICNALE-AS2R}}
\newcommand{\Biaffine}{\mbox{\sc Biaf}}

\newcommand{\rellabel}[1]{\textit{#1}}
\definecolor{SuperLightGray}{gray}{0.9}
\newcommand{\MAR}[1]{\ifthenelse{\equal{#1}{}}{MAR}{MAR$^{\text{#1}}$}}

\title{Multi-Task and Multi-Corpora Training Strategies to Enhance Argumentative Sentence Linking Performance}

\author{
    Jan Wira Gotama Putra$^\dag$, Simone Teufel$^{\ddag\dag}$, Takenobu Tokunaga$^\dag$ \\
    $^\dag$Tokyo Institute of Technology, Japan \\ 
    $^\ddag$University of Cambridge, United Kingdom \\ 
    {\small \tt \{gotama.w.aa@m, take@c\}.titech.ac.jp, simone.teufel@cl.cam.ac.uk} \\
}

\begin{document}
\maketitle
\begin{abstract}

Argumentative structure prediction aims to establish links between textual units and label the relationship between them, forming a structured representation for a given input text. The former task, linking, has been identified by earlier works as particularly challenging, as it requires finding the most appropriate structure out of a very large search space of possible link combinations. 
In this paper, we improve a state-of-the-art linking model by using multi-task and multi-corpora training strategies. 
Our auxiliary tasks help the model to learn the role of each sentence in the argumentative structure. Combining multi-corpora training with a selective sampling strategy increases the training data size while ensuring that the model still learns the desired target distribution well.
Experiments on essays written by English-as-a-foreign-language learners show that both strategies significantly improve the model's performance; for instance, we observe a 15.8\% increase in the F1-macro for individual link predictions.  

\end{abstract}

\section{Introduction}
\label{sec:intro}

Argument mining (AM) is an emerging area that addresses the automatic analysis of argumentation. Many recent studies commonly try to tackle two major tasks \citep{10.1162/coli_a_00364}.
The first of these is \textit{argumentative component identification}, in which argumentative units (ACs) and non-argumentative components (non-ACs) including their boundaries are determined. 
ACs can be further classified according to their role in argumentation, e.g., \textit{major claim},\footnote{The major claim is the statement expressing the writer's view on the discussion topic; also called \textit{main stance} or \textit{main claim}.} \textit{claim} and \textit{premise} \citep{Stab:2017:PAS:3160785.3160790}. The second task is called \textit{argumentative structure prediction}, which first establishes links from source to target ACs (this is called the \textit{linking} task) and then labels the relationship between them, for instance using the \rellabel{support} and \rellabel{attack} relation labels (this is called the \textit{relation labelling} task) \citep{Stab:2017:PAS:3160785.3160790}.

\begin{figure}[ht]
    \centering
    \includegraphics[width=1.0\linewidth]{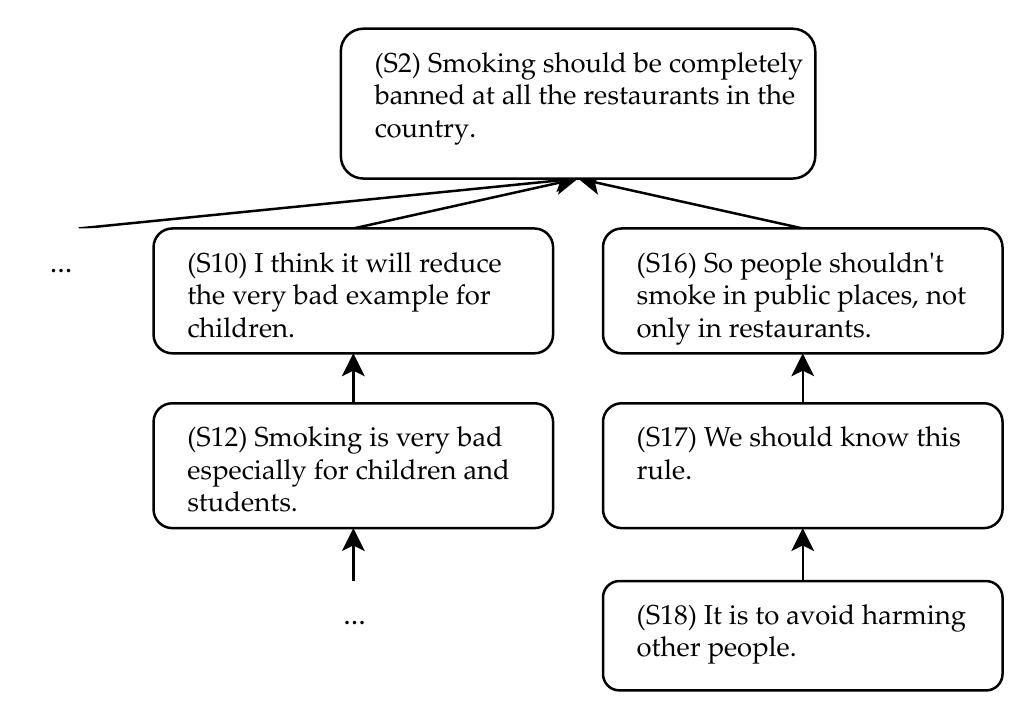}
    \caption{Illustration of linking task, using part of an essay discussing the topic ``\textit{Smoking should be banned at all restaurants in the country}."}
    \label{fig:annotation_example}
\end{figure}

The linking task has been identified by earlier works as particularly challenging \citep{Lippi:2016:AMS:2909066.2850417,Cabrio-Villata-2018,10.1162/coli_a_00364}.
There are many possible combinations of links between textual units, and a linking model has to find the most proper structure out of a very large search space.
Another typical challenge in AM is the size of annotated corpora \cite{schulz-etal-2018-multi}. Corpus construction is a complex and time-consuming process; it also often requires a team of expert annotators. Existing corpora in AM are relatively ``small" compared with more established fields, such as machine translation or document classification. This hinders training AM models when using a supervised machine learning framework. 

In this paper, we perform the linking task for essays written by English-as-a-foreign-language (EFL) learners. Given an essay, we identify links between sentences, forming a tree-structured representation of argumentation in the text. Figure~\ref{fig:annotation_example} illustrates the task. 
Our contributions are twofold. First, we propose two structural-modelling related auxiliary tasks to train our model in a multi-task learning (MTL) fashion. Past studies commonly executed several AM tasks or using discourse and rhetorical auxiliary tasks in the MTL setting (cf. Section~\ref{sec:related_work}). In comparison with other works, our auxiliary tasks are advantageous in that they do not require any additional annotation. Second, we combine multi-corpora training with a selective sampling strategy to increase the training data size. We explore the use of mixed training data of in-domain EFL and out-domain\footnote{Past studies used the term \textit{domain} in a broad context. It has at least five different senses: text genre, text quality, annotation scheme, dataset and topic (or prompt). In the rest of this paper, we use the specific meaning whenever possible.} student-essay corpora that were annotated using different schemes and of different quality, while ensuring that the model still learns the properties of the target in-domain data well.

The choice of EFL texts in this study aims to contribute to a less attended area. In AM, it is common to use well-written texts by proficient authors \citep[e.g.,][]{ashley-legal-argument,PeldszusStede-ECA:16}. However, student texts often suffer from many problems because they are still in the learning phase. Even more, EFL texts are also less coherent and less lexically rich, and exhibit less natural lexical choices and collocations \citep{silva-1993,rabinovich-etal-2016-similarities}. There are more non-native English speakers \cite{fujiwara-teaching-grammar-2018}, and yet, to the best of our knowledge, only one preceding study in AM concerned the EFL genre \citep{putra-etal-2021-parsing}. The codes accompanying this paper are publicly available.\footnote{\url{https://github.com/wiragotama/ArgMin2021}}

\section{Related Work}
\label{sec:related_work}

\subsection{Argumentative Structure Prediction}
\label{ssec:RW_ASP}

A variety of formulations have been proposed for the linking task. Traditional approaches formulated it as a pairwise classification task, predicting whether an argumentative link exists between a given pair of ACs \citep{stab-gurevych-2014-identifying}. A further post-processing step can also be performed to combine the local predictions into an optimised global structure, e.g., using the minimum-spanning-tree algorithm \citep{peldszus-stede-2015-joint}.
Recent studies proposed a more global approach instead, considering the entire input context.
For instance, \citet{potash-etal-2017-heres} formulated the linking task as a sequence prediction problem. They jointly performed AC classification and AC linking at once, assuming that the segmentation and AC vs non-AC categorisation have been pre-completed. 
They experimented on the \textit{microtext corpus} \citep{PeldszusStede-ECA:16} and the \textit{persuasive essay corpus} (PEC, \citet{Stab:2017:PAS:3160785.3160790}).

\newcite{eger-etal-2017-neural} formulated argumentative structure parsing in three ways: as relation extraction, as sequence tagging and as dependency parsing tasks. They defined a BIO tagging scheme and performed end-to-end parsing at token-level, executing all subtasks (i.e., segmentation, unit type classification, linking and relation labelling) at once. 
\citet{ye-teufel-2021-eacl} also performed end-to-end parsing at the token-level. They proposed a more efficient representation for the dependency structure of arguments, and achieved the state-of-the-art performance for component and relation identifications on the PEC using a biaffine attention model \citep{dozat-manning-2017-ICLR}. 

The biaffine attention model was originally designed to parse token-to-token dependency, but \citet{morio-etal-2020-towards} extended it to parse proposition (segment) level dependency. Their model dealt with graph-structured arguments in the Cornell eRulemaking corpus \citep{park-cardie-2018-corpus}. 
Using the same architecture, \citet{putra-etal-2021-parsing} parsed tree-structured EFL essays in the \ICNALEargs{} corpus \citep{ICNALE-general,ICNALE-edited,wira_nle_2021,putra-etal-2021-parsing}. In tree-structured argumentation, it is common for groups of sentences about the same sub-topic to operate as a unit, forming a sub-tree (sub-argument). \citet{putra-etal-2021-parsing} found that their linking model has problems in constructing sub-trees, that is, it splits a group of sentences that should belong together into separate sub-arguments (sub-trees) or, conversely, groups together sentences that do not belong together into the same sub-trees.

\subsection{Low-Resource and Cross-Domain Argument Mining}
\label{ssec:RW_data_augmentation}

Several approaches have been applied to alleviate the data sparsity problem in AM. \citet{al-khatib-etal-2016-distant-supervision} used a distant supervision technique to acquire a huge amount of data without explicit annotation. \citet{accuosto-saggion-2019-transferring} pre-trained a discourse parsing model and then fine-tuned it on AM tasks. \citet{lauscher-etal-2018-investigating} investigated the MTL setup of argumentative component identification and rhetorical classification tasks. \citet{schulz-etal-2018-multi} performed a cross-genre argumentative component identification. They employed a sequence tagger model with a shared representation but different prediction layers for each genre. 

Data augmentation can also be applied to mitigate the data sparsity problem. This aims to increase the amount of training data without directly collecting more data \citep{liu-data-augmentation-2020,Feng-data-augmentation-survey-2021}. 
A relatively straightforward strategy is to use multiple corpora when training models. For example, \citet{chu-etal-2017-empirical} proposed a \textit{mixed fine-tuning} approach for machine translation; they trained a model on an out-genre corpus and then fine-tune it on a dataset that is a mix of the target-genre and out-genre corpora. 
However, the use of multiple corpora of different genres is challenging in AM because argumentation is often modelled differently across genres \citep{Lippi:2016:AMS:2909066.2850417,10.1162/coli_a_00364}. \citet{daxenberger-etal-2017-essence} found that training a claim identification model with mixed-genre corpora only perform as good as training on each specific corpus. The use of data augmentation may cause the distributional shift problem as well, where the augmented data alter the target distribution that should be learned by the model \citep{Feng-data-augmentation-survey-2021}.

\citet{putra-etal-2021-parsing} provided some insights on the possible conditions for multi corpora training by experimenting using texts of different quality. They trained argumentative structure prediction models on EFL essays (in-domain) and parallel improved version of the texts (out-domain), and evaluated on the EFL texts. Their cross-domain system attained 94\% accuracy of end-to-end in-domain performance. This signals the potential to use corpora of different text quality altogether to train a parsing model, as long as the text genre stays the same. However, mixed-quality corpora training has yet to be tried in practice. 

In recent years, there has also been a growing interest towards more generic AM models. For example, \citet{stab-etal-2018-cross} proposed a simple annotation scheme for argument retrieval that is applicable in heterogeneous sources and can be performed by untrained annotators. \citet{cocarascu-2020} explored various deep learning architectures and features that work well across various datasets for the relation labelling task. 
They also provided a comparison of using contextualised and non-contextualised embeddings for the task. 
On the other hand, \citet{Dolz-2021-rel-labelling} compared the performance of transformer-based language models on the cross-topic relation labelling task.

\section{Dataset}
\label{sec:dataset}

Our target texts are sourced from the \ICNALEargs{}, a corpus of 434 essays written by Asian college students with intermediate proficiency.\footnote{\small{\url{https://www.gsk.or.jp/en/catalog/gsk2021-a}}. The approximated CEFR level of the essay authors are A2 (94 essays), B1 (253) and B2 (87).} There are 6,021 sentences in total with 13.9 sentences on average per essay. 
To the best of our knowledge, this is the only currently publicly-available AM corpus focusing on EFL texts.
The corpus is annotated at sentence-level, that is, a sentence corresponds to one argumentative unit. The corpus differentiates sentences as ACs (5,799 sents.) and non-ACs (222 sents.), without further classification of AC types. Links are established between ACs to form tree structure (avg. depth of 4.3, root at depth 0 in the corpus), where the \textit{major claim} acts as the root. Four relation labels are employed to label the links: \rellabel{support}, \rellabel{attack}, \rellabel{detail} and \rellabel{restatement}.

\section{Proposed Method}
\label{sec:proposed_method}

In this work, we experiment on the linking task. Given an entire essay of $N$ sentences as input, $s_1, ..., s_N$, a linking model outputs the distance $d_1, ..., d_N$ between each sentence $s_i$ to its target. For instance, if a sentence is connected to its preceding sentence, the distance is $d=-1$. We consider those sentences that have no explicitly annotated outgoing links as linked to themselves ($d=0$); this concerns major claims (roots) and non-ACs. We do not consider labelling the links with their relationships in this paper.

\begin{figure}[ht]
    \centering
    \includegraphics[width=1.0\linewidth]{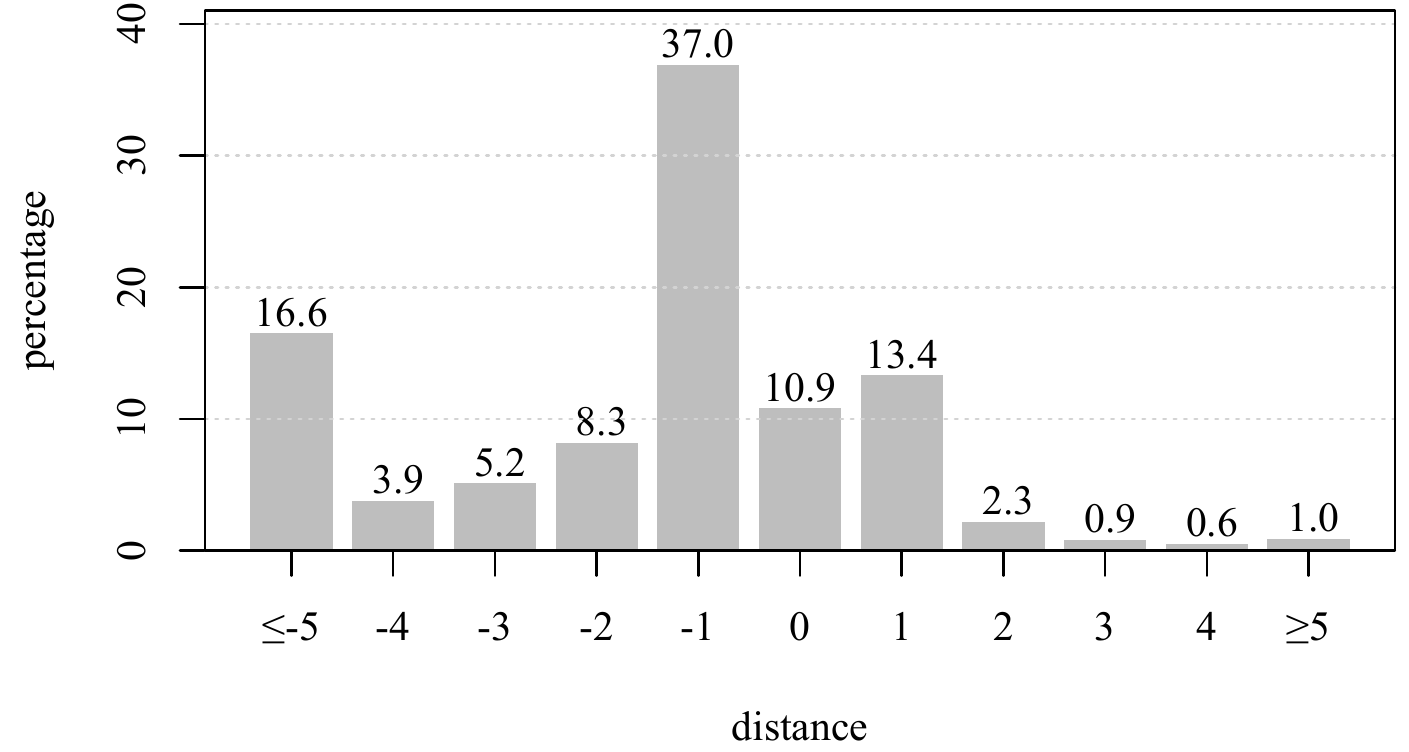}
    \caption{Distribution of distance (in percent) between related sentences in the \ICNALEargs{} corpus.}
    \label{fig:distance_distribution}
\end{figure}

Figure~\ref{fig:distance_distribution} shows the distance distribution between the source and target sentences in the \ICNALEargs{} corpus, ranging $[-26, ..., +15]$. Adjacent links predominate ($50.4\%$). Short distance links ($2\leq |d| \leq 4$) make up $21.2\%$ of the total. Backward long distance links at $d\leq-5$ are $16.6\%$, whereas forward long distance links are rare ($1.0\%$). Self-loops make up 10.9\% of the total.
Following recent advances in AM, we approach the linking task as a sentence-to-sentence dependency parsing.

\subsection{Baseline}
\label{ssec:baseline}

\begin{figure}[ht]
    \centering
    \includegraphics[width=1.0\linewidth]{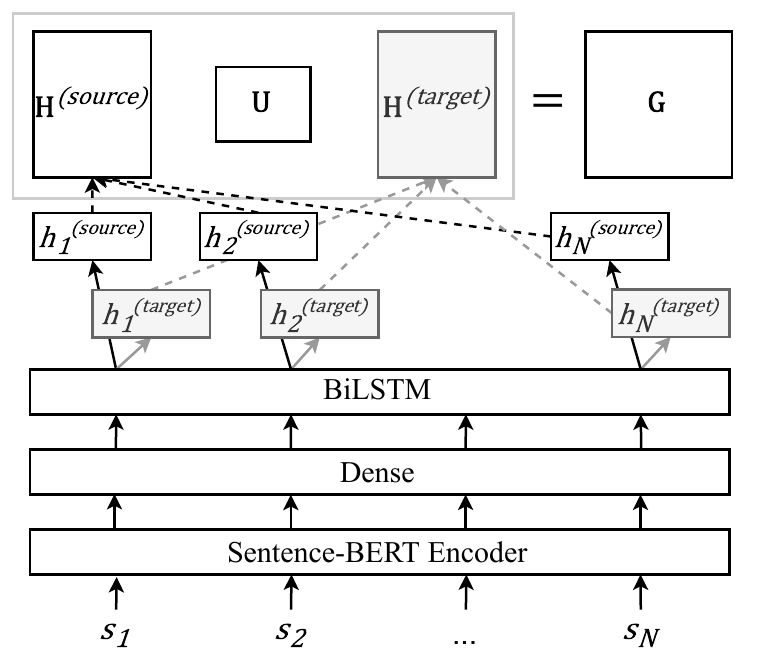}
    \caption{Biaffine attention model (\Biaffine{}).}
    \label{fig:BiaffineBase}
\end{figure}

We use \citeauthor{putra-etal-2021-parsing}'s (\citeyear{putra-etal-2021-parsing}) architecture as the state-of-the-art baseline (``\Biaffine{}" model, cf. Figure~\ref{fig:BiaffineBase}). In this architecture, input sentences are first encoded into their respective sentence embeddings using sentence-BERT encoder \citep{reimers-gurevych-2019-SBERT}.
The resulting sentence embeddings are then passed into a dense layer for dimensionality reduction. The results are then fed into a bidirectional long-short-term memory network (BiLSTM) layer (\#$stack=3$) to produce contextual sentence representations. These representations are then passed into two different dense layers to encode the corresponding sentence when it acts as a source $\left( h^{(source)} \right)$ or target $\left( h^{(target)} \right)$ in a relation.

Finally, a biaffine transformation \citep{dozat-manning-2017-ICLR} is applied to all source and target representations to produce the final output matrix $\mathrm{\mathbf{G}} \in \mathbb{R}^{N\times N}$, where each cell $g_{i,j}$ represents the probability (or score) of the source sentence $s_i$ pointing to $s_j$. Equation~(\ref{eq:biaffine}) and (\ref{eq:biaffine_result}) show the biaffine transformation ($f$), where $\mathrm{\mathbf{U}}$ and $\mathbf{\mathrm{W}}$ are weight matrices and $b$ is bias. 
\begin{align}
    f(x_1, x_2) &= \begin{aligned}[t]
            {x_1}^{\mathrm{T}} \mathrm{\mathbf{U}} x_2 + \mathrm{\mathbf{W}} (x_1 \oplus x_2) + b
    \label{eq:biaffine}
    \end{aligned}\\
    g_{i,j} &= \begin{aligned}[t]
        f \left( h_i^{(source)}, h_j^{(target)} \right)
    \label{eq:biaffine_result}
    \end{aligned}
\end{align}
The Chu-Liu-Edmonds algorithm \citep{Chu1965OnTS, edmonds-1967} is applied to create a minimum spanning tree out of the matrix $\mathrm{\mathbf{G}}$. The links in the tree are then converted into distance predictions between source and target sentences for evaluation purpose.

\begin{figure*}[ht]
    \centering
    \includegraphics[width=0.90\linewidth]{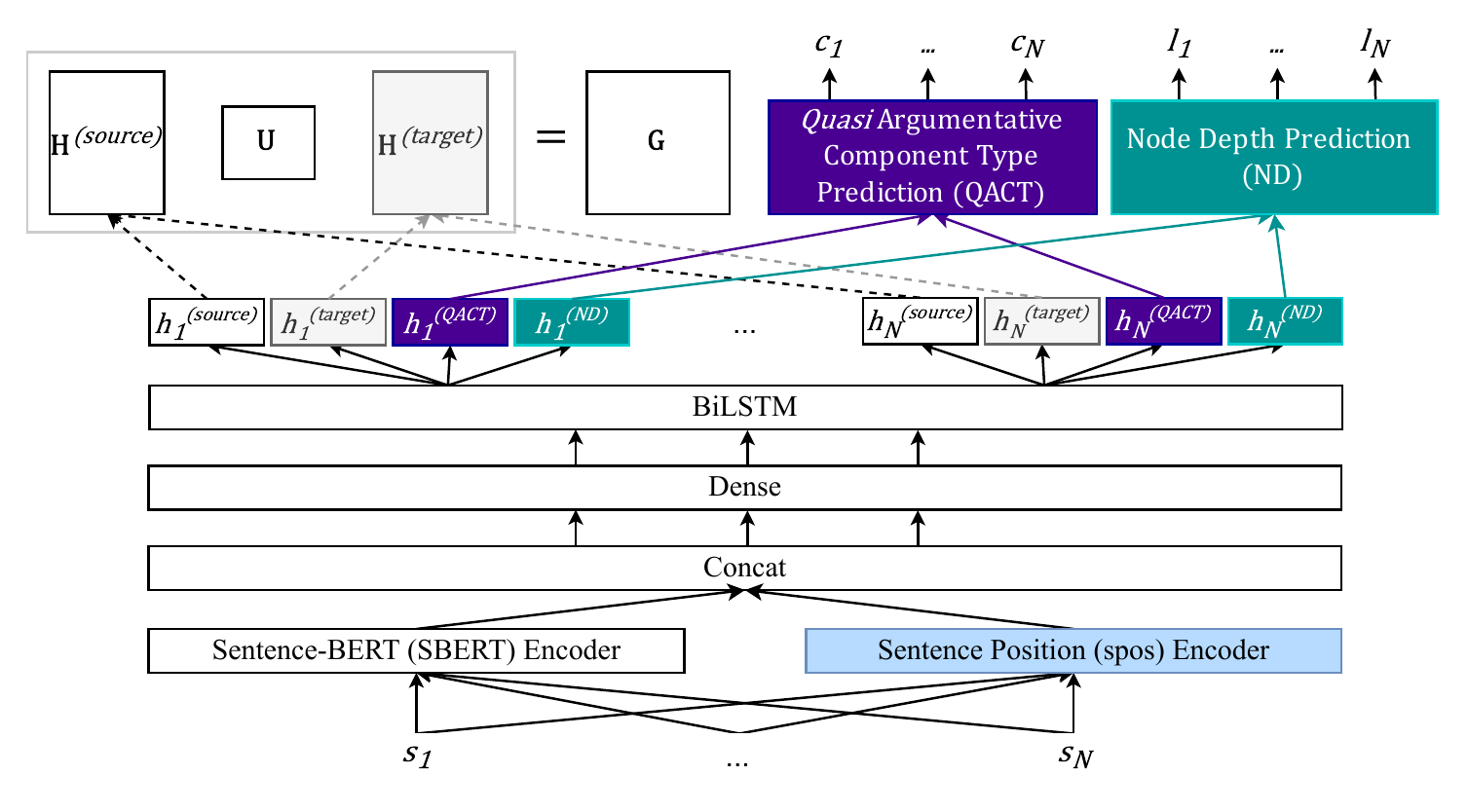}
    \caption{Our proposed extension for the \Biaffine{} model. Newly added modules are coloured.}
    \label{fig:Biaffine_MTL}
\end{figure*}

\subsection{Multi-Task Learning with Structural Signal}
\label{ssec:MTL}

We propose to extend the \Biaffine{} model in an MTL setup using two novel structural-modelling-related auxiliary tasks. 

The first auxiliary task is a \textit{quasi} argumentative component type (QACT) prediction. \ICNALEargs{} corpus does not assign AC types per se, but we can compile the following four sentence types from the tree typology:
\begin{itemize}[noitemsep]
    \item \textit{major claim (root)}: only incoming links,
    \item \textit{AC (non-leaf)}: both outgoing and incoming links,
    \item \textit{AC (leaf)}: only outgoing links and
    \item \textit{non-AC}: neither incoming nor outgoing links.
\end{itemize}
For example, S2 is the major claim of Figure~\ref{fig:annotation_example}, S10 is AC (non-leaf) and S17 is AC (leaf). The QACT prediction task should help the linking model to learn the role of sentences in the argumentative structure, as well as the property of links for each sentence. This auxiliary task improved a sequence-tagger based linking model in \citet{putra-etal-2021-parsing}, but it has not been applied to the \Biaffine{} model yet. 

The second auxiliary task concerns node depth (ND) prediction. There are six depth categories employed in this paper: depth 0 to depth 4, and depth 5+. The argumentative structure in \ICNALEargs{} corpus is hierarchical, and there is no relations between nodes (sentences) at the same depth. The ND prediction task should help the model to learn the placement of sentences in the hierarchy and guide where each sentence should point at, that is, sentences at depth $X$ point at sentences at depth $X-1$. 

We also propose to use sentence position (spos) embedding as an input feature because it has been proved to be useful in other studies \citep[e.g.,][]{song-etal-2020-discourse}. The sentence position encoding is calculated by dividing a sentence position by the essay length. When using this feature, sentence embeddings and spos encoding are concatenated before being passed to the first dense layer. Figure~\ref{fig:Biaffine_MTL} illustrates our proposed architecture.

\subsection{Multi-Corpora Training}
\label{ssec:multi_copora_training}

In this paper, we also consider the use of multi-corpora training strategy, employing essays in the PEC as additional training data. Both \ICNALEargs{} and PEC provide argumentative essays written by students (in English) and represents argumentative structures as trees. However, there is no information on the essay authors proficiency (L1 or L2) nor the observed quality of the PEC. Hence, these two corpora might be of different quality. The writing prompts of \ICNALEargs{} and PEC essays are also different.

There are two settings when training using multiple corpora. The first is to use the entire 402 essays in the PEC on top of \ICNALEargs{} train set (``[P+I]" setting). However, PEC and \ICNALEargs{} are different in terms of length and annotation scheme. The PEC essays have 18.2 sentences on average (15.1 ACs and 3.1 non-ACs), while \ICNALEargs{} essays have 13.9 sentences on average (13.4 ACs and 0.5 non-ACs). The difference in non-ACs proportion between these corpora is likely caused by the difference in the set of relation labels employed. The essays in PEC were annotated using two relation types: \rellabel{support} and \rellabel{attack}, while \ICNALEargs{} additionally employed \rellabel{detail} and \rellabel{restatement}. Hence, there is arguably more information in the \ICNALEargs{} corpus. Particularly, some sentences that should have been annotated as non-ACs (hence not linked to other sentences) in the PEC's scheme might have been annotated as ACs in the \ICNALEargs{}'s scheme by using the additional relations. 

Due to the differences in annotation scheme and statistical properties, there is a possibility that the model might not properly learn the distribution of \ICNALEargs{} in the [P+I] setting. Therefore, we also propose a second \textit{selective sampling} (``[SS]" setting) strategy to account for the differences between these two corpora. In this setting, we only use PEC essays that are somewhat similar to those of \ICNALEargs{} considering the following two criteria. First, we only use PEC essays having 17 sentences at maximum (\ICNALEargs{} avg. 13.9 + 3.3 SD). Second, the employed PEC essays contain two non-ACs at maximum (\ICNALEargs{} avg. 0.5 + 0.9 SD).
There are 110 remaining PEC essays after the selective sampling procedure.

\begin{figure}[ht]
\centering
    \exampletext{
    $_\mathrm{(S1)}$[\textit{To conclude}, \textbf{art could play an active role in improving the quality of people's lives,}] $_\mathrm{(S2)}$[\textit{but I think that} \textbf{governments should attach heavier weight to other social issues such as education and housing needs}] $_\mathrm{(S3)}$[\textit{because} \textbf{those are the most essential ways enable to make people a decent life.}]
    }
\caption{Illustration of conversion of PEC's segment-level annotation to the sentence-level (essay075). Annotated AC segments are written in bold.}
\label{fig:PEC_to_sentence_ann}
\end{figure}

Note that \ICNALEargs{} was annotated at the sentence level while PEC was annotated at the segment level. Therefore, we convert the PEC annotation to the sentence level, following the strategy described by \citet{song-etal-2020-discourse}. If a sentence contains only one AC, we use the whole sentence as an AC; if a sentence contains two or more ACs, we split it into multiple sentences while including the preceding tokens into each AC. Figure~\ref{fig:PEC_to_sentence_ann} illustrates the splitting procedure, where a sentence containing three ACs is split into three sentences.

\section{Structural Metric}
\label{sec:substructure_metric}

In this paper, we do not evaluate the model performance only on individual link predictions but also analyse the structural properties of the outputs, giving more insights into the models' ability to learn different aspects of the argumentative structure. We also investigate whether our proposed strategies improve the model performance concerning the grouping of related sentences into the same sub-tree, a challenge to linking models previously (manually) observed in \citet{putra-etal-2021-parsing} (cf. Section~\ref{ssec:RW_ASP}). A metric is needed to quantify the improvement on this aspect, and we propose to use a novel structural metric called \textbf{\MAR{dSet}}, that was developed by \citet{wira_nle_2021}.\footnote{MAR stands for \textbf{M}ean \textbf{A}greement in \textbf{R}ecall. Descendant set (dSet) denotes the unit of analysis.}

\begin{figure}[ht]
    \centering
    \includegraphics[width=1.0\linewidth]{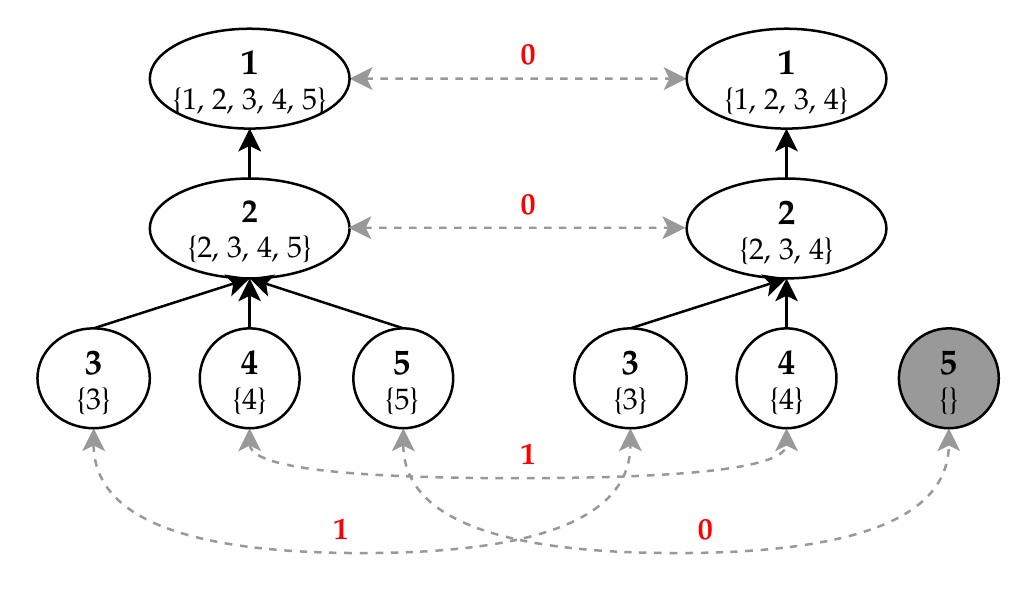}
    \caption{Example of descendant set matching between structure $A$ (left) and $B$ (right). Matching scores are written in red. Grey nodes represent non-AC.}
    \label{fig:substructure_matching}
\end{figure}

Given two structures $A$ and $B$ (i.e., predicted and gold structures), we can quantify their similarity based on the presence of common substructures. We first define a \textit{descendant set} (dSet) of a node~$X$ as the set consisting the node~$X$ itself and its descendants. Figure~\ref{fig:substructure_matching} shows examples of dSets (brackets given below the node ID).\footnote{In the implementation, we use sentence position in the input text as node ID.} For example, the dSet of node-2 in annotation $A$ of Figure~\ref{fig:substructure_matching} is $\{2,3,4,5\}$. Two corresponding nodes in $A$ and $B$ are required to have identical dSets in order to score a value of 1. For example, the matching score for node-2 between annotation $A$ and $B$ in Figure~\ref{fig:substructure_matching} is 0, while the matching score for node-3 is 1. Non-AC nodes are counted as a match if they are deemed non-argumentative in both structures. 

With this formulation, we can get a vector $v$ representing the matching scores for all nodes in the two structures. For example, $v=[0,0,1,1,0]$ for Figure~\ref{fig:substructure_matching}. The similarity score between two structures are then calculated as \MAR{dSet}$=\frac{\sum_{i=1}^{N}v_i}{N}$, where $N$ denotes the number of nodes; The \MAR{dSet} score for Figure~\ref{fig:substructure_matching} is 0.4.

\section{Experiment and Discussion}
\label{sec:experimental_results}

\subsection{Experimental Setting}
\label{ssec:experimental_setting}

In this experiment, the \ICNALEargs{} corpus is split into 80\% train set (347 essays , 4,841 sents.) and 20\% test set (87 essays, 1,180 sents.). We use the same splits as \citet{putra-etal-2021-parsing}. The number of training essays for [P+I] and [SS] settings are 749 (12,162 sents.) and 457 (6,418 sents.), respectively. We run the experiment for 20 times\footnote{Twenty experiments were repeated on the same dataset splits to account for random initialisation in neural networks.} and report the average performance. Statistical testing, whenever possible, is conducted using the permutation test \citep{permutation-test} on the performance scores of the 20 runs with a significance level of $\alpha=.05$. Here, we also perform an ablation study. Appendix~\ref{sec:appendix_implementation} shows some implementation notes.

\subsection{Quantitative Analysis}
\label{ssec:quantitative_analysis}

Table~\ref{tab:individual_link_results} shows the experimental results on individual link predictions and \MAR{dSet}. Training the \Biaffine{} model using the QACT task improved the performance over the baseline, particularly in terms of F1-macro (non significant difference). In addition, using both QACT+ND auxiliary tasks significantly improved the performance over the baseline in terms of F1-macro. This signals that the proposed MTL setting benefits model performance.

\begingroup
\setlength{\tabcolsep}{4pt}
\begin{table}[ht]
    \centering
    \small
	{\begin{tabular}{lrrr}
		\toprule
		\textbf{Model} & \textbf{Acc.} & \textbf{F1-macro} & \textbf{\MAR{dset}} \\
		\midrule
		\textit{Baseline} \\
		\Biaffine           & .471 & .323 & .419 \\
		\midrule
		\textit{MTL} \\
		\Biaffine+QACT      & .473 & .333 & 422 \\
		\Biaffine+QACT+ND   & .472 & .338 & 423 \\
		\midrule
		\textit{Spos}\\
		\Biaffine+QACT+spos     & .472 & .327 & .421 \\
		\Biaffine+QACT+ND+spos   & \ul{.475} & .336 & 426 \\
		\midrule
		\textit{Multi-corpora Training} \\
		\Biaffine+QACT+ND [P+I]   & .468 & \ul{.360} & \textbf{.455} \\
		\Biaffine+QACT+ND [SS]     & \textbf{.489}$^\dag$ &  \textbf{.374}$^\dag$ & \ul{.452} \\
		\bottomrule
	\end{tabular}}
	\caption{Results of individual link predictions (Acc. and F1-macro) and \MAR{dSet}. The best result is shown in \textbf{bold-face}. The $\dag$ symbol indicates that the difference to the second-best result (underlined) is significant.}
	\label{tab:individual_link_results}
\end{table}
\endgroup

We next look at how spos encoding affects the model performance.
Introducing spos to the \Biaffine{}+QACT+ND model improves accuracy and \MAR{dSet}.
However, the difference is not significant. Similarly, the difference between \Biaffine{}+QACT+spos and \Biaffine{}+QACT is not significant.
There are two possible explanations for this phenomenon. First, studies in \textit{contrastive rhetoric} found that non-native speakers tend to structure and organise their texts differently from native speakers \citep{Kaplan1966,silva-1993}. Some students might have used the writing customs and rhetorical strategies of their first language instead of using the common English writing patterns. For example, 
Asian students sometimes put a claim after its reason, 
which is not common in Anglo-Saxon cultures \citep{silva-1993,johns-1986}. We observe that some texts in the \ICNALEargs{} corpus are written in the ``claim-support" structure but some are written in the ``support-claim" structure. Sentences on the same topic are sometimes distantly separated as well. These inconsistencies might have negated the effect of the spos encoding. Second, the output of the \Biaffine{} model is a graph $\mathrm{\mathbf{G}}$ which considers the directed links between all pairs of sentences. The spos feature might not affect the biaffine transformation much in this context. 

The models trained using multiple corpora attain the best performance for individual link and substructure (MAR) predictions. \Biaffine{}+QACT+ND [P+I] achieves the best performance of .455 in terms of \MAR{dSet}, and \Biaffine{}+QACT+ND [SS] achieves the best performance of .489 and .374 in terms of accuracy and F1-macro. The [SS] model also achieves the second-best performance of .452 in terms of \MAR{dSet}. These improvements are significant over the baseline and the \Biaffine{}+QACT+ND model. Note that when using the [SS] setting, the model performance is consistently improved with regard to all metrics, while the accuracy of the [P+I] model is lower than the baseline. In general, the [SS] model attains a better performance compared with the [P+I] model despite fewer training instances. This means that when training a model using multiple corpora, it is essential to consider training instances having the same properties as our goal. Simply having more training data does not guarantee improvement. 


\begin{figure}[ht]
    \centering
    \includegraphics[width=1.0\linewidth]{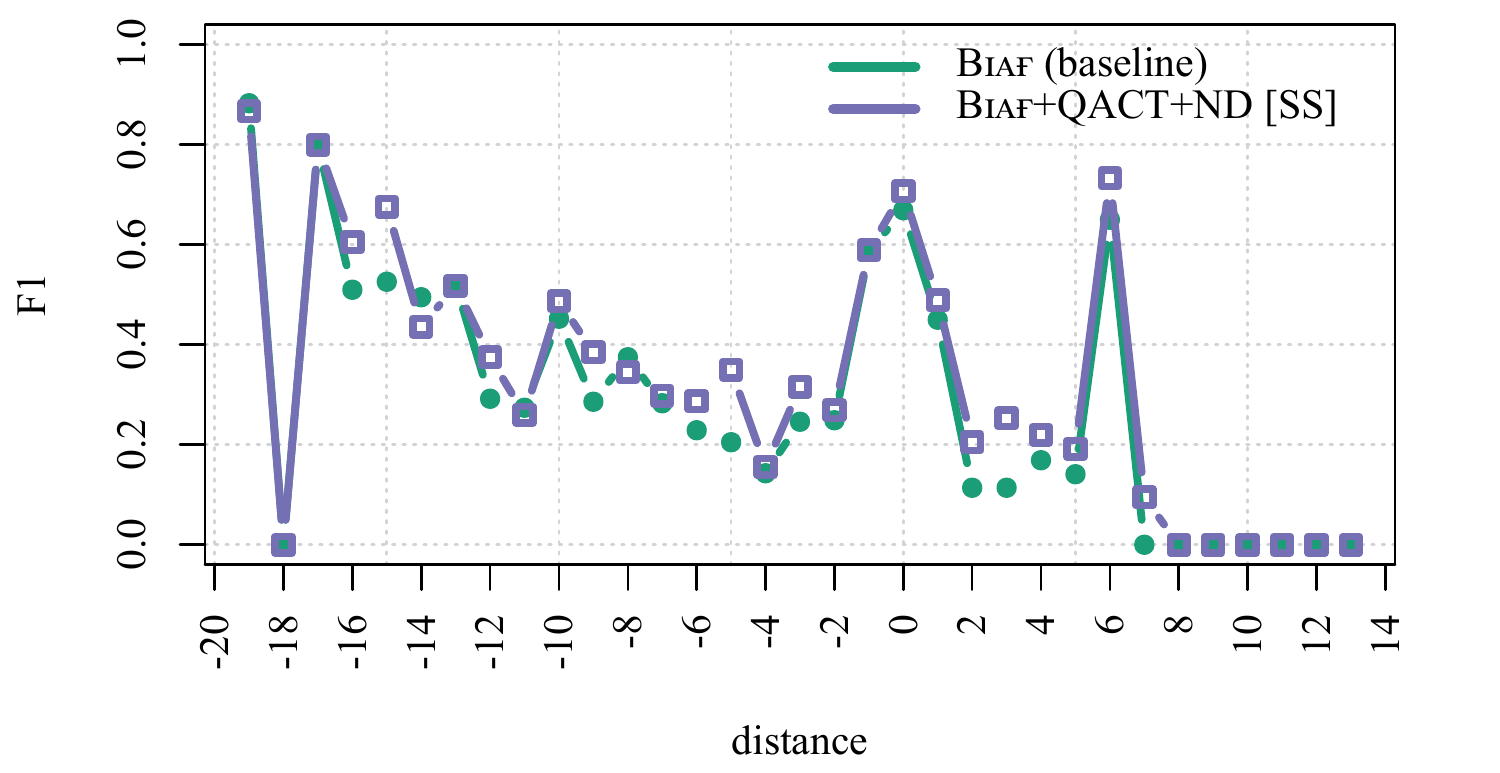}
    \caption{Model performance across distances.}
    \label{fig:F1_dist_linking_rich}
\end{figure}

\begin{figure}[ht]
    \centering
    \includegraphics[width=1.0\linewidth]{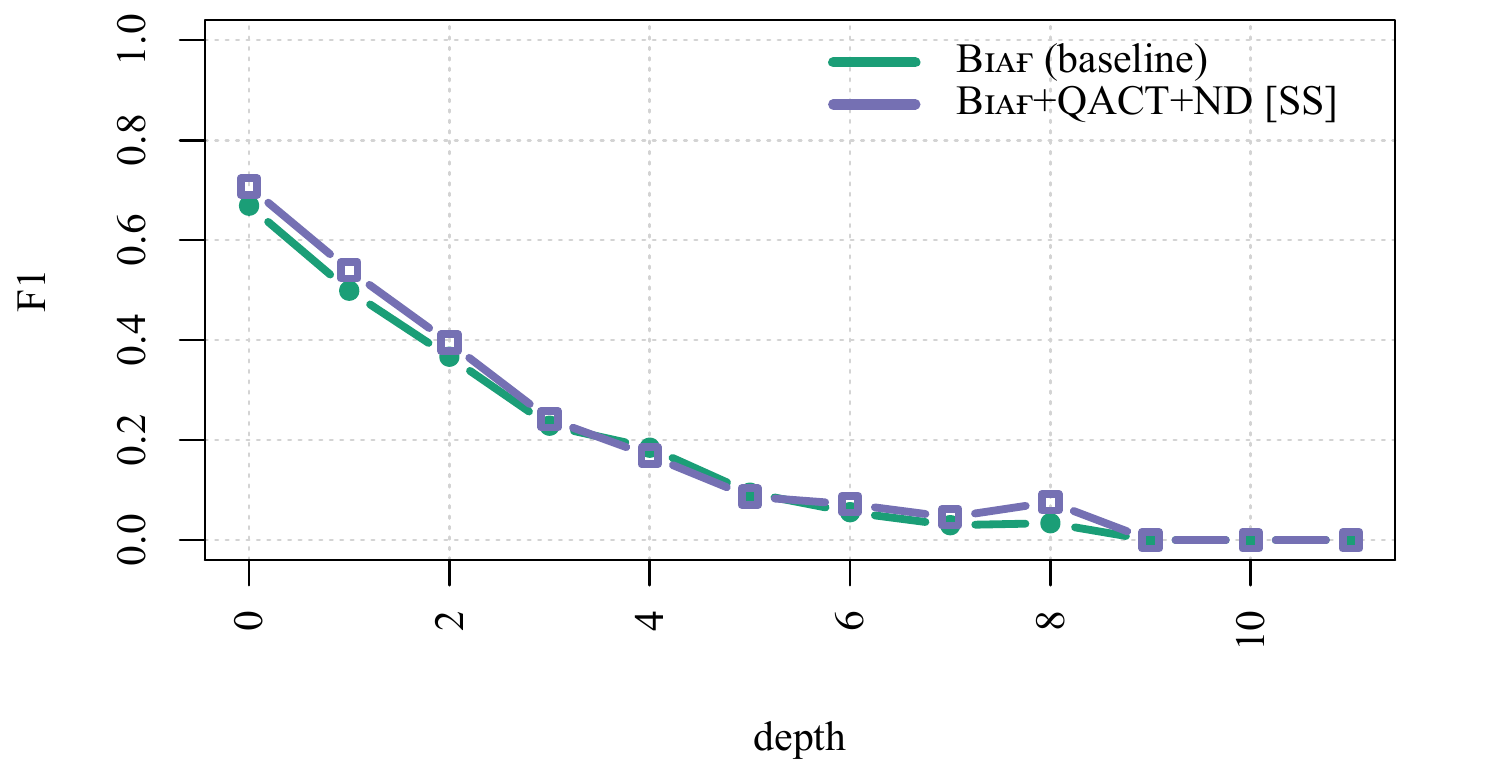}
    \caption{Model performance across depths.}
    \label{fig:F1_depth_linking_rich}
\end{figure}

To gain deeper insights into the individual link prediction improvement brought by the [SS] model over the baseline, Figure~\ref{fig:F1_dist_linking_rich} shows F1 score per target linking distance. \Biaffine{}+QACT+ND [SS] is better than the baseline model, particularly at predicting short-distance links ($2\leq |d| \leq 4$, avg. F1 = .24 vs .17), which is the weakest range of the baseline model. However, this is still the weakest range even for the [SS] model.

Figure~\ref{fig:F1_depth_linking_rich} shows the performance across depths, that is, whether the model places each node at the proper depth 
in the predicted structure. \Biaffine{}+QACT+ND [SS] performs better than the baseline particularly in [0,3] and [6,8] ranges. This plot indicates that the model performance declines as one moves further down the tree.


\begingroup
\begin{table*}[htpb]
    \centering
    \small
	{\begin{tabular}{lrrrrr}
		\toprule
		\textbf{Model} & \textbf{Major Claim} & \textbf{AC (non-leaf)} & \textbf{AC (leaf)} & \textbf{Non-AC} & \textbf{F1-macro} \\
		\midrule
		\textit{Baseline} \\
		\Biaffine           & .730 & \ul{.639} & .599 & .437 & .601 \\ 
		\midrule
		\textit{MTL} \\
		\Biaffine+QACT      &  .739 & \ul{.639} & .601 & .453 & .608 \\
		\Biaffine+QACT+ND   &  .734 & .636 & .601 & .454 & .606 \\
		\midrule
		\textit{Spos} \\
		\Biaffine+QACT+spos     &  .725 & \textbf{.641} & .602 & .438 & .602 \\
		\Biaffine+QACT+ND+spos  &  .738 & .638 & .603 & \ul{.460} & \ul{.610} \\ 
		\midrule
		\textit{Multi-corpora Training} \\
		\Biaffine+QACT+ND [P+I]   &  \ul{748} & .606 & \textbf{.634}$^\dag$ & .420 & .602 \\
		\Biaffine+QACT+ND [SS]     & .\textbf{767}$^\dag$ & .633 & \ul{.628} & \textbf{.462} & \textbf{.622}$^\dag$ \\
		\bottomrule
	\end{tabular}}
	\caption{Results for the QACT prediction. Node labels are automatically identified from the predicted topology in the main task. This table shows F1 score per node label and F1-macro. \textbf{Bold-face}, $\dag$ and underline as above.}
	\label{tab:rich_linking_QACT_results}
\end{table*}
\endgroup

We next look at the models' ability to predict the role of each node (QACT task) based on the predicted hierarchical structures. Table~\ref{tab:rich_linking_QACT_results} shows the results. The MTL models perform better compared with the baseline model. Both \Biaffine{}+QACT and \Biaffine{}+QACT+ND achieve a significant improvement over the baseline in terms of non-AC prediction and F1-macro. This reconfirms that both of our proposed MTL tasks are useful to improve linking performance.
Similar to the previous result on individual links, the spos feature does not provide much improvement.
We notice that the \Biaffine{}+QACT+ND [P+I] model performs worse than the baseline in some aspects, particularly in terms of AC (non-leaf) and non-AC predictions. On the other hand, \Biaffine{}+QACT+ ND [SS] consistently achieves better performance compared with the baseline model. It attains the highest score of .622 in F1-macro, which is a significant improvement over other configurations.
This confirms our hypothesis that the difference between annotation schemes and statistical properties between PEC and \ICNALEargs{} affects the model in terms of the distribution learned. The proposed selective sampling strategy helps to alleviate this problem. 


\begin{table}[ht]
    \centering
    \small
	{\begin{tabular}{lrrr}
		\toprule
		\textbf{Model} & \textbf{Average Depth} & \textbf{Leaf Ratio} \\
		\midrule
		\textit{Dataset} \\
		\ICNALEargs{}       & 4.3$\pm$1.4 & .439$\pm$.11 \\
		PEC [all]           & 2.8$\pm$.63 & .540$\pm$.09 \\
		PEC [after SS]      & 2.7$\pm$.53 & .565$\pm$.08 \\
		\midrule
		\textit{Baseline} \\
		\Biaffine           & 5.1 & .404 \\
		\midrule
		\textit{MTL} \\
		\Biaffine+QACT      & 5.1 & .410 \\
		\Biaffine+QACT+ND   &  5.0 & \ul{.418} \\
		\midrule
		\textit{Spos} \\
		\Biaffine+QACT+spos     & 5.2 & .407 \\
		\Biaffine+QACT+ND+spos  & 5.0 & .412 \\
		\midrule
		\textit{Multi-corpora Training} \\
		\Biaffine+QACT+ND [P+I]   &  \ul{4.1} & .486 \\
		\Biaffine+QACT+ND [SS]    & \textbf{4.5}$^\dag$ & \textbf{.446}$^\dag$ \\
		\bottomrule
	\end{tabular}}
	\caption{Structural-output qualities of linking models. \textbf{Bold-face}, $\dag$ and underline as above.}
	\label{tab:rich_linking_structural_shape}
\end{table}

We also analyse the overall shape of the predicted structures by all models, as shown in Table~\ref{tab:rich_linking_structural_shape}. The gold standard trees in \ICNALEargs{} have a particular shape, expressed as the average depth of 4.3 ($\mathrm{SD}=1.4$) and the leaf ratio of .439 ($\mathrm{SD}=0.11$).
The baseline model tends to produce trees that are deeper and narrower than the \ICNALEargs{} gold trees. Our MTL auxiliary tasks help to improve the leaf ratio to become closer to the gold standard (significant difference between \Biaffine{}+QACT+ND and \Biaffine{}), while spos embedding does not provide additional improvement. When we introduce the multi-corpora training strategy, the predicted structures become shallower and wider compared with the baseline, particularly when using the [P+I] setting. We believe this is due to the shallower trees in the PEC compared with the \ICNALEargs{} (i.e., the distributional shift problem). 
However, this is less of an issue for the [SS] model as it produces the most similar structure to the \ICNALEargs{} essays. 

We conclude \Biaffine{}+QACT+ND [SS] as the best model in this experiment, achieving the new state-of-the-art performance for the linking task. It consistently performs better compared with other configurations across all evaluation aspects. 


\subsection{Qualitative Analysis}
\label{qualitative_analysis}

\begin{figure}[ht]
    \centering
    \includegraphics[width=1.0\linewidth]{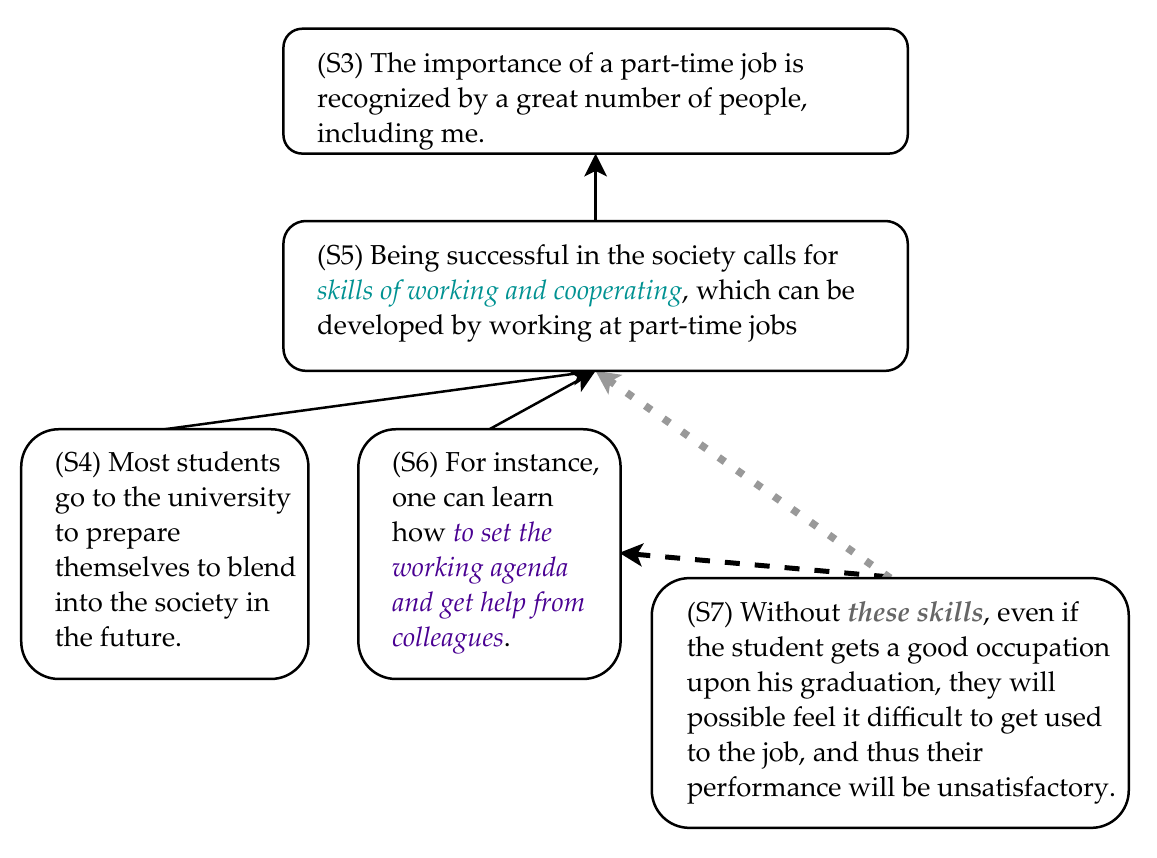}
    \caption{An excerpt of prediction for essay ``W\_HKG\_PTJ0\_029\_B2\_0\_EDIT". The essay was written in response to a prompt ``\textit{It is important for college students to have a part-time job}." The dashed line illustrates an erroneous link, while the dotted line (grey) illustrates the corresponding gold link.}
    \label{fig:output_example_gold_pred_combined}
\end{figure}

We conducted a qualitative analysis on some random outputs of our best model, namely \Biaffine+QACT+ND [SS]. Some prediction errors are likely caused by the inability of the model to resolve anaphora. Figure~\ref{fig:output_example_gold_pred_combined} shows an output example. The model successfully groups sentences S5--S7 as a single sub-argument, as the gold annotation does. However, it connects S7 to S6, while the gold annotation connects S7 to S5 (with the \textit{support} label). The link from S7 to S5 is the correct interpretation because the statement in S7 can be broadly applied to ``\textit{skills of working and cooperating}," not only to some specific instantiations mentioned in S6 (``\textit{set the working agenda and get help from colleagues}"). 
A possible direction to alleviate this problem is to equip the model with a better ability to resolve anaphora. Another direction is to perform an anaphora resolution step before parsing the argumentative structure.

\begin{figure}[ht]
    \centering
    \includegraphics[width=1.0\linewidth]{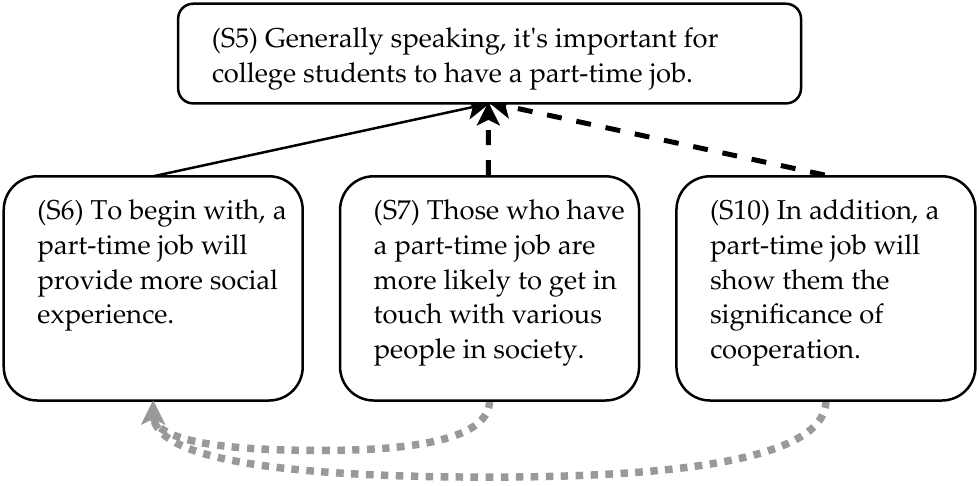}
    \caption{An excerpt of prediction for essay ``W\_CHN\_PTJ0\_242\_B2\_0\_EDIT". The essay was written in response to a prompt ``\textit{It is important for college students to have a part-time job}." Dashed lines and dotted lines as above.}
    \label{fig:output_example_shape_only}
\end{figure}

Another source of error is the difficulty in judging whether two statements argue at the same conceptual level (hence the same depth in the tree). Figure~\ref{fig:output_example_shape_only} shows an illustration. S7 and S10 are exemplifications of S6, and therefore, they should be placed lower than S6. However, the model recognises S6, S7 and S10 arguing at the same conceptual level. The exemplifications are probably not obvious for the model because S7 is not accompanied by a discourse marker (e.g., ``\textit{for example}"). In general, EFL essays are often challenging to process due to improper use (can be excessive or limited too) of discourse markers.

\section{Conclusion}
\label{sec:conclusion}

In this paper, we presented a study on the argumentative linking task; given an essay as input, a linking model outputs the argumentative structure in a tree representation. We conducted experiments using an EFL corpus, namely \ICNALEargs{}. We extended the state-of-the-art biaffine attention model using a novel set of structural auxiliary tasks in the multi-task learning setup. Additionally, we also proposed a multi-corpora training strategy using the PEC to increase training instances. It has to be noted that simply increasing the training data size does not guarantee improvement. We need to ensure that the model still learns the desired target distribution as well. To this end, we filtered PEC essays using a selective sampling technique. These two strategies provided useful supervision signals to the biaffine model and significantly improved its performance. The F1-macro for individual link predictions was boosted to .374 from .323. Our strategies also improved the model performance on the structural aspects, achieving the \MAR{dset} of .452 from .419, and the F1-macro of .622 from .601 for the QACT task. 

A possible future direction is to evaluate whether our proposed methods are beneficial for more complex discourses, such as scientific articles. It is also necessary to provide an even richer supervision signal to improve the linking performance. One possibility is to pre-train the model on the anaphora resolution task.

\section*{Acknowledgements}

This work was partially supported by Tokyo Tech World Research Hub Initiative (WRHI), JSPS KAKENHI grant number 20J13239 and Support Centre for Advanced Telecommunication Technology Research. We would like to thank anonymous reviewers for their useful and detailed feedback.

\bibliography{references}
\bibliographystyle{acl_natbib}

\appendix

\section{Implementation Notes}
\label{sec:appendix_implementation}

\paragraph{Hidden Units and Model Implementation} 512 units for the first dense layer, 256 LSTM unit, and 256 units for the second dense layer after the BiLSTM layer. Dropout is applied between all layers. For a fair comparison, we follow \citet{putra-etal-2021-parsing} in using the sentence-BERT encoder fine-tuned on the NLI dataset (``bert-base-nli-mean-tokens", \url{https://github.com/UKPLab/sentence-transformers}). All models are trained using Adam optimiser~\cite{adam-optimizer}, and implemented in PyTorch \citep{pytorch} and AllenNLP \citep{allennlp}.

\paragraph{Loss} The MTL loss is defined as $L = \sum_{t} \frac{1}{2\sigma_{t}^2}L_{t} + \mathrm{ln}(\sigma_{t})$, where the loss $L_{t}$ of each task $t$ is dynamically weighted, controlled by a learnable parameter $\sigma_t$. The loss for the main task (sentence linking) is computed using max-margin criterion, while losses for auxiliary tasks are computed using cross-entropy. 

\end{document}